\documentclass[12pt,conference,onecolumn,a4paper]{IEEEtran}
\IEEEoverridecommandlockouts
\usepackage{cite}
\usepackage{amsmath,amssymb,amsfonts}
\usepackage{url}
\usepackage{diagbox}
\usepackage{graphicx}
\usepackage{textcomp}
\usepackage{xcolor}
\usepackage{listings}
\usepackage{hyperref}
\usepackage{verbatim}
\usepackage{geometry} 

\usepackage{algcompatible}
\usepackage{algpseudocode}
\usepackage{algorithm}
\usepackage{algorithmicx}

\def\BibTeX{{\rm B\kern-.05em{\sc i\kern-.025em b}\kern-.08em
    T\kern-.1667em\lower.7ex\hbox{E}\kern-.125emX}}

\begin{document}
\newgeometry {top=30mm,left=30mm, right= 30mm,bottom =30mm} 

\newtheorem{definition}{Definition}
\newtheorem{example}{Example}

\title{Forest Mixing: investigating the impact of multiple search trees and a shared refinements pool on ontology learning
}

\author{\IEEEauthorblockN{Marco Pop-Mihali and Adrian Groza}
\IEEEauthorblockA{\textit{Department of Computer Science} \\
\textit{Technical University of Cluj-Napoca, Romania}\\
Cluj-Napoca, Romania \\
Pop.Si.Marco@student.utcluj.ro, Adrian.Groza@cs.utcluj.ro}
}

\maketitle

\begin{abstract}
We aim at development white-box machine learning algorithms.
We focus here on algorithms for learning axioms in description logic. 
We extend the Class Expression Learning for Ontology Engineering (CELOE) algorithm contained in the DL-Learner tool. 
The approach uses multiple search trees and a shared pool of refinements in order to split the search space in smaller subspaces. 
We introduce the conjunction operation of best class expressions from each tree, keeping the results which give the most information. 
The aim is to foster exploration from a diverse set of starting classes and to streamline the process of finding class expressions in ontologies. 
The current implementation and settings indicated that the Forest Mixing approach did not outperform the traditional CELOE. 
Despite these results, the conceptual proposal brought forward by this approach may stimulate future improvements in class expression finding in ontologies. 
\end{abstract}


\section{\textbf{Introduction}}
Machine learning models are being deployed across diverse sectors, from predicting outcomes in business, guiding decision-making in finance, to advancing diagnostics and treatment planning in medicine. However, a significant challenge of these models is their "black box" nature. 
Complex models built with deep learning networks are not easily interpretable, lacking understanding of how they derive their predictions or decisions. 
This lack of transparency can pose serious issues, especially when they are applied to critical areas where interpretability and explainability are needed.

In contrast, "white box" models offer insights into the decision-making process, indicating the influence each feature has on the output. 
They present a more transparent approach for predicting outcomes, but these advances often come with a performance trade-off. 
Such models may not deliver performance on par with Large Language Models (LLMs), or they might require more time and resources to offer similar outputs.

As building blocks for the Semantic Web, ontologies can be used for data storage, relations among these data, reasoning, or as a background knowledge source for machine learning algorithms. 
A specific task is finding class expressions from ontologies and examples, an area that might be approaced by inductive logic programming (ILP).

We present a novel approach to inductive logic programming, in which we modify the state-of-the-art algorithm CELOE~\cite{b2}. 
Our Forest Mixing approach aims to improve the process of finding class expressions from ontologies and traversing large search spaces, offering a potentially more efficient solution to this type of problems.

\section{\textbf{Related Work}}
The learning algorithm proposed here belongs to the larger field of Inductive Logic Programming (ILP). 
ILP represents a fusion between inductive learning and logic programming, aiming to derive hypotheses from observations and to create new knowledge from experience. 

ALEPH (A Learning Engine for Proposing Hypotheses) is a tool that operates within the domain of Inductive Logic Programming (ILP)~\cite{b3}. 
ALEPH formulates hypotheses based on a given set of positive and negative examples and a body of background knowledge. It utilizes a 'set covering' loop and applies a hill-climbing search strategy within the hypothesis space. This approach is governed by a refinement operator, facilitating the exploration of the hypothesis space. The versatility of ALEPH, demonstrated by its adaptable parameters and settings, enables it to handle a wide array of logic programming tasks, making it a significant tool in the field of ILP. Notably, ALEPH has found successful applications across various sectors, such as bioinformatics and natural language processing~\cite{b3}.

DL-Learner (Description Logic Learner) is a framework for supervised machine learning in Semantic Web and other structured knowledge domains. 
Using refinement operators, the tool is designed for learning concepts within Description Logics (DLs), including other related formalisms such as OWL and RDF(S). 
Among its multiple learning algorithms, the CELOE begins with a broad concept (e.g., "owl:Thing" in OWL) and incrementally refines it, aiming to discover the most specific concepts that satisfy predefined quality criteria based on given positive and negative examples. 
The algorithm leverages a heuristic search, which enables efficient handling of large knowledge bases by removing the need for exhaustive searches~\cite{b1,b2}.
We rely on the modular design of the DL-Learner tool, which allows easy extension of the CELOE algorithm and easy reuse of its components like the Refinement Operators.

Learning ontologies have been also explored with Relational Concept Analysis~\cite{b9}, semantic role labelling~\cite{b20}, or Large Language Models (LLMs)~\cite{b10}. 
Role labbeling has been use to fill the gap between natural language expressions and ontology concepts or properties~\cite{b20}.
The LLMs are fine tuned to translate from natural language to OWL functional syntax. 
The generated translations can be manually validated by the human agent through a plugin for the Protoge ontology editor.  

\section{\textbf{Theoretical instrumentation}}
We briefly introduce here some theoretical notions like: ontologies, description logics and refinement operators. 

Ontologies are a key component in semantic web technologies and knowledge representation systems. 
They provide a structured framework of concepts and their relationships, facilitating more effective information retrieval, data integration, and reasoning.
They contain clases (i.e. concepts, sets), relations among these classes (that can have some properties like reflexivity, transitivity, symmetry), and individuals (instances of concepts). 

Description Logic (DL) is a formal language utilized for knowledge representation, often deployed in Semantic Web and ontologies for class expression and querying. 
DL exhibits a balance between expressivity and computational efficiency. 
The expressivity of DL stems from operators used for creating complex classes, as outlined in Table~\ref{table:DL_Operators}~\cite{b5, b6}.
In DL, ontologies are formalised using a collection of concepts (classes), roles (relationships), and individuals. 
By reasoning in DL, one can perform automatic consistency checking, or maintaining the integrity of the knowledge base when introducing new facts~\cite{b6}. 

\begin{table}
\caption{Description Logic Operators}
\begin{center}
\label{table:DL_Operators}
\begin{tabular}{ll}
\hline
\textbf{DL Operator} & \textbf{Description} \\
\hline
$\top$ (top) & Special concept with every individual as an instance \\
$\bot$ (bottom) & Empty concept \\
$\sqcap$ (and) & Intersection or conjunction of concepts \\
$\sqcup$ (or) & Union or disjunction of concepts \\
$\neg$ (not) & Negation or complement of concepts \\
$\forall$ (for all) & Universal restriction \\
$\exists$ (exists) & Existential restriction \\
$\sqsubseteq$ (is-a) & Inclusion of the concept \\
$\equiv$ (equivalent) & Equivalence of concepts \\
$ \dot =$ (definition) & Definition of the concept \\
$:$ (assertion) & Assertion of the concept \\
\hline
\end{tabular}
\end{center}
\end{table}

\begin{definition}
A refinement operator $\rho$ is a mapping from a concept $C$ to a set of concepts, such that: $\rho: \mathcal{C} \rightarrow {\mathcal{C}_1, \mathcal{C}_2, ..., \mathcal{C}_n}$, where each $\mathcal{C}_i$ represents a hypothesis.
\end{definition}
 Refinement operators can be classified into two main types: downward refinement operators and upward refinement operators.

\begin{definition}
A downward refinement operator, denoted as $\rho^{\downarrow}(\mathcal{C})$, transforms a concept $C$ into a set of more specific concepts $\mathcal{C}_1, \mathcal{C}_2, ..., \mathcal{C}_n$, where each $\mathcal{C}_i \subseteq \mathcal{C}$ for all $i = 1, 2, ..., n$.
\end{definition}
\begin{example}
Let the current class expression $C =Bird$. 
Applying a downward refinement operator, a more specific class expression is obtained, as $Bird\ \sqcap\ \exists hasFeature.Fly$, describing birds that fly. 
The new expression describe a smaller set of individuals. Similarly, when the $\neg$ operator is applied, one can obtain the expression $Bird\ \sqcap\ \neg Aquatic$, describing birds that are not aquatic. 
\end{example}

\begin{definition}
An upward refinement operator, denoted as $\rho^{\uparrow}(\mathcal{C})$, transforms a concept $\mathcal{C}$ into a set of more general concepts $\mathcal{C}_1, \mathcal{C}_2, ..., \mathcal{C}_n$, where each $\mathcal{C}_i \supseteq \mathcal{C}$ for all $i = 1, 2, ..., n$.
\end{definition}

\begin{example}
Let the initial expression $C =Birds\ \sqcap\ Carnivore$. 
Applying an upward refinement operator on $C$, a more general expression is obtained, that is $Bird$, corresponding to a larger set of individuals. 
\end{example}

Refinement operators are used to generate and test hypotheses during learning. 
By applying these operators to concept learning, they facilitate navigation through the large space of possible hypotheses~\cite{b7}.

\section{\textbf{Forest Mixing Approach}}
We start by analysing aspects of the state-of-the-art CELOE approach that can be improved. 
Building on these observations, we formalise the novel Forest Mixing approach (FM) for ontology learning. 


\subsection{Potential Advantages of FMA}
In both Forest Mixing approach and Random Forest algorithms, the search space is divided among several smaller trees. However, this division does not amount to a strict partition in either of the methods. 
Random Forests train each  tree on subsets of overlapping data and features.
In FM, each tree navigates a subset of the search space, but these subsets are not mutually exclusive. Trees might delve into similar or even the same parts of the search space. 
A crucial difference emerges in the way overlaps are addressed in these algorithms. For Random Forests, overlapping can be beneficial, while for FM, redundancies arising from multiple trees generating identical class expressions can increase computational costs. Despite these contrasts, the central concept of FM draws inspiration from the Random Forest's mechanism.

Though CELOE~\cite{b2} stands as the state-of-the-art in ontology-based hypothesis search, there exist scenarios where its performance might be improved. 
Within the scope of CELOE, and hypothesis searching in general, the most computationally demanding operation is the refinement process. This operation can induce an exponential growth in the number of nodes (concepts or class expressions) to be examined. Therefore, an efficient algorithm in our context should ideally minimize the number of refinements required to find the best hypothesis. 
Note that both CELOE and FM approach provide the functionality to set initial concepts. 
Setting the initial concepts with the help of user's knowledge triggers a reduction in the search space. 
We hypothesize two cases where the Forest Mixing approach could offer more efficiency than CELOE.

First, FM can exhibit higher efficiency compared to CELOE, particularly when users have prior knowledge of the data and can suggest starting classes within relevant subspaces. 
For example, consider non-disjoint classes such as $Employee$ and $Student$, where individuals could be part of both classes. 
If the target concept is for instance 
\begin{equation}
Student \sqcap Employee \sqcap \exists attendsCourse .EveningCourse    
\end{equation}
representing individuals enrolled in a university and working there who also attend evening courses, there can be useful hypotheses within both $Employee$ and $Student$ subspaces. 
In this case, FM's parallel exploration can expedite the process by simultaneously investigating both paths, potentially discovering a suitable hypothesis faster than sequentially exploring one subspace after the other as CELOE would do.

Second, FM potentially outperforms CELOE in cases involving disjunctions in the target concept. 
Disjunctions pose a challenge for most Inductive Logic Programming (ILP) techniques, including CELOE, due to the prevalent use of downward refinement operators. These operators primarily generate conjunctions, not disjunctions. 
For instance, consider the target concept 
\begin{equation}
(Student \sqcup Employee) \sqcap \exists attends AICourse    
\end{equation}
representing individuals who are either students or employees and attend an $AICourse$. CELOE, in this case, might need to explore a vast search space exhaustively. 
FM can address this efficiently by assigning different starting classes or computing them, thus facilitating parallel exploration in different relevant subspaces. 
While FM approach may not directly find the exact class expression, it can swiftly uncover simpler, separate class expressions such as:
\begin{eqnarray}
Student \sqcap \exists attends.AICourse \\
Employee \sqcap \exists attends.AICourse
\end{eqnarray}

\subsection{Designing the FM algorithm}
The FMA commences by selecting an initial class or classes as the starting point. 
This selection is a strategic move aimed at reducing the search space, consequently increasing the efficiency of the algorithm. 
The criterion for choosing a class is its ability to contain all the positive examples, symbolized by $\mathcal{C}$. 
Such a class can then be refined or specialized without any loss of positive examples, as it ensures a full positive coverage $PosCov$:

\begin{equation}
PosCov(ce) = \frac{|ce_{pos}(E)|}{|E_{pos}|} \label{eq:pozcov}
\end{equation}

Here, $PosCov(ce)$ is the positive coverage of the class expression $ce$. 
$ce_{pos}(E)$ represents the set of positive examples covered by $ce$, 
and $E_{pos}$ is the set of all positive examples. 
Therefore, a class with a $PosCov$ value of 1.0 signifies that all positive examples are encapsulated within that class. 
This ensures that the search space is efficiently minimized from the outset, providing an optimized starting point for further refinement and specialization. 
The selection of best nodes is described by Algorithm 1. 

\begin{center}
\includegraphics[width=\textwidth]{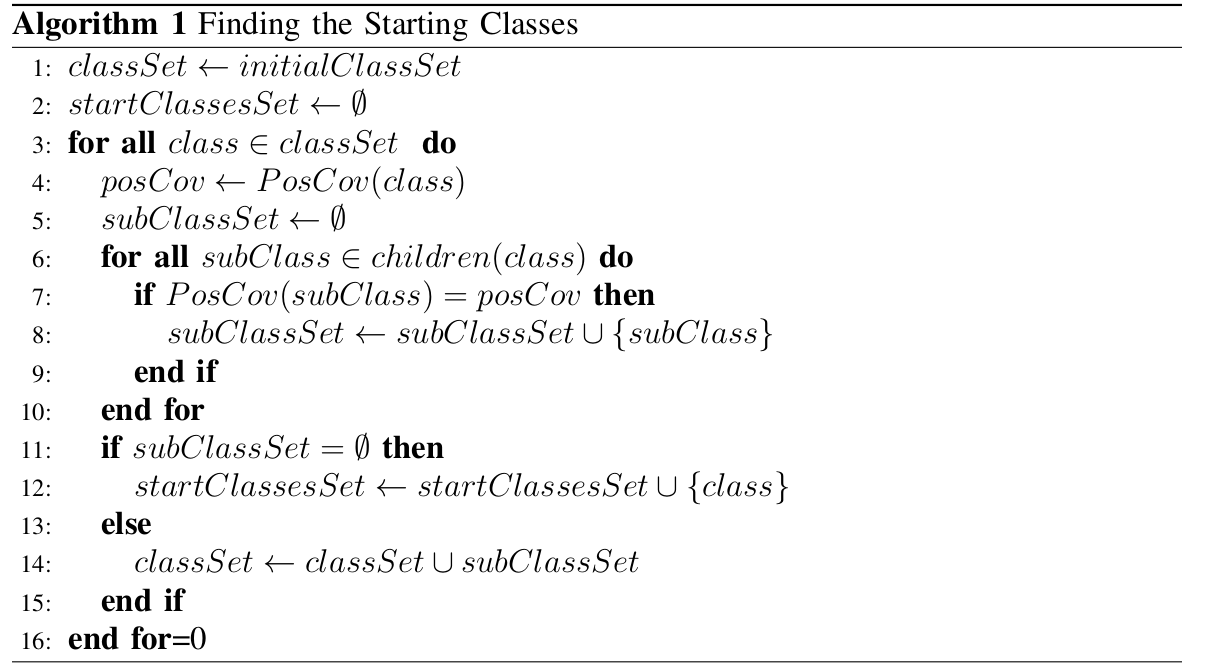} 
\end{center}

The Algorithm 1 starts by applying a strategy similar of CELOE, where a search tree is generated, the best nodes identified, and refined. 
However, FM approach introduces the following enhancement - instead of maintaining a single search tree, it manages multiple trees, and all the refined expressions derived are kept in a shared pool. 
Each tree is permitted to draw a maximum number of expressions from the shared refinements, consequently  promoting an efficient and diversified exploration of the search space.

The process of adding refinements to the shared pool is governed by specific conditions. First, the algorithm maintains a record of the best nodes from previous trees and also 
the refinements added by the current node, 
Second, the current expression is checked against this list. 
If the current expression and any of the previous bests do not share a class in common, a conjunction of them is computed. 
This conjunction is only added to the shared poll when all the classes are distinct, aiming to maximize the class expressions which bring new information, and which do not have multiple identical classes. 

The complexity of a node (i.e. class expression) is measured as the length of the expression, thereby a conjunction of two expressions might be overly complex. 
To avoid this case, the length of the resulting expressing is added to the shared pool only when it doesn't exceed a threshold set using a FM parameter. 
The conjunction selection process is shown in the Algorithm 2 
The rest of the algorithm, refining nodes, selecting best nodes is very similar to CELOE~\cite{b2}. 

\begin{center}
\includegraphics[width=\textwidth]{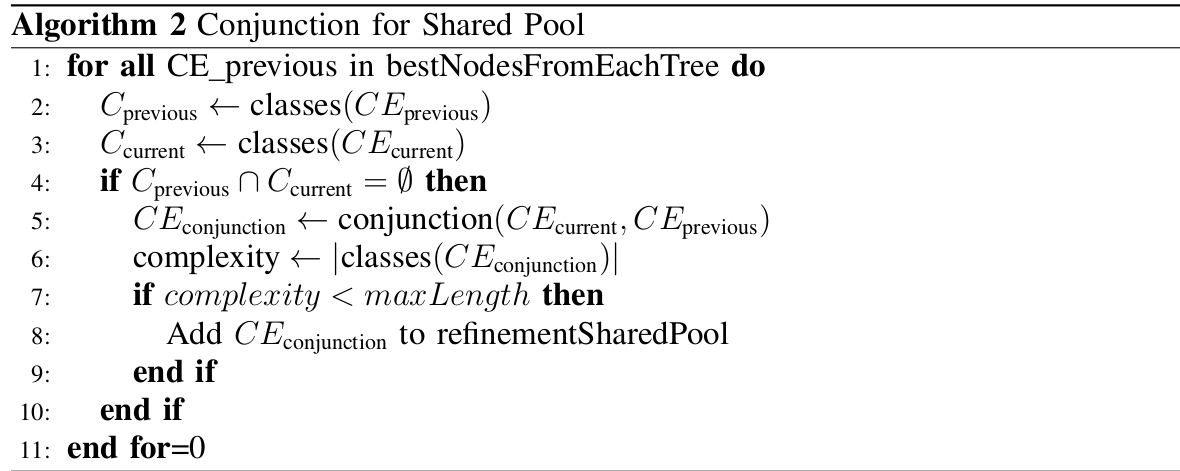} 
\end{center}

\begin{figure}
\centerline{\includegraphics[width=0.6\textwidth]{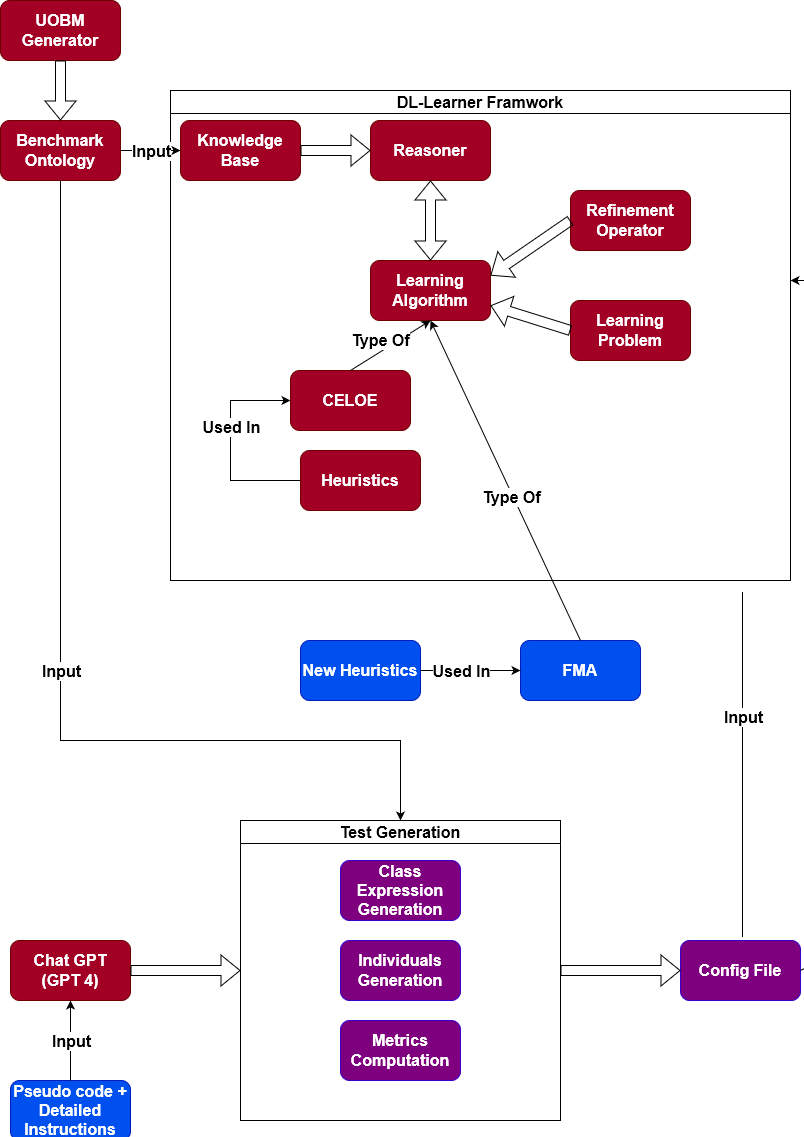}}
\caption{System Architecture}
\label{fig:SysArch}
\end{figure}

FM represents an extension of the CELOE approach (see Figure~\ref{fig:SysArch}). 
To evaluate the proposed FM, we rely on the University Ontology Benchmark (UOBM) generator. 
The UOBM generator outputs scalable and realistic ontologies, tailored for benchmarking ontology-based systems~\cite{b8}.  

\begin{figure*}
\centerline{\includegraphics[width=1.1\textwidth]{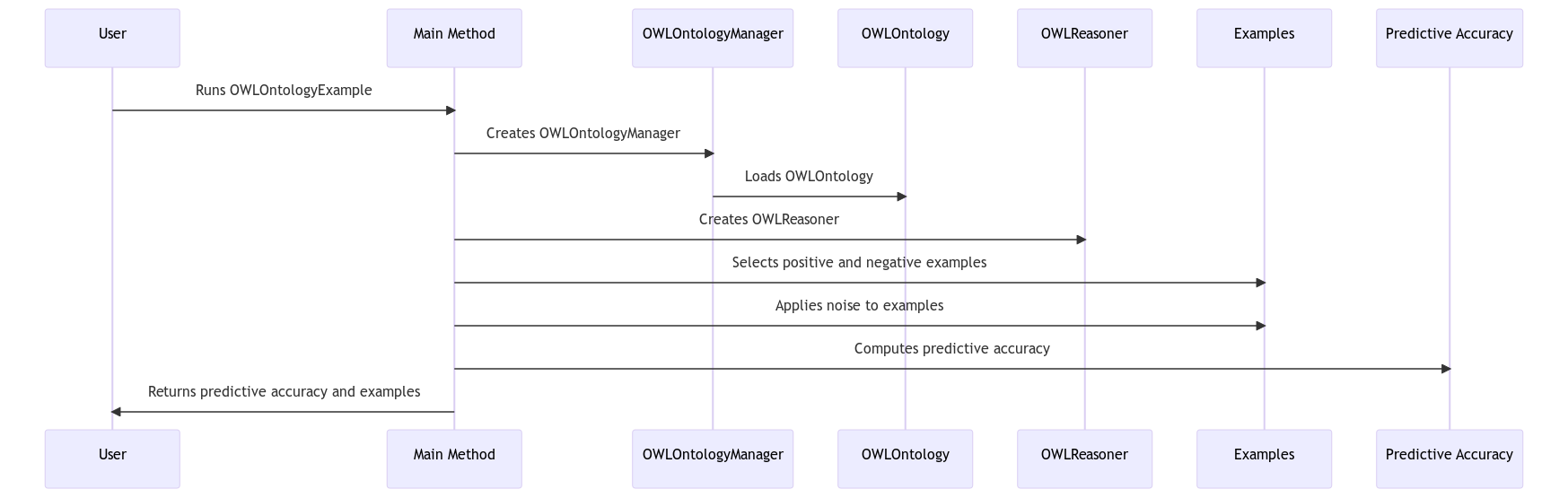}}
\caption{Generating testing ontologies}
\label{fig:seqTestGen}
\end{figure*}

Ontologies alone do not suffice for generating a complete test. 
In the case of a specific ontology generated with the UOBM generator, we require a class expression (also called ground truth or target) along with two sets of individuals: one belonging to the class and the other not. 
To handle this, we designed the Algorithm 3 
that finds a suitable class expression and individuals in the given ontology. 
The corresponding diagram flow appears in Figure~\ref{fig:seqTestGen}. 
To implement the proposed algorithm we rely on GPT-4.  
Human implementation was used in a few selected places where the code generated did not capture the correct logic steps. 
This is why in Figure~\ref{fig:SysArch} GPT-4 is represented as an external system (i.e. colored red). 
The test generation modules are colored purple, which means they are the result of computer generation and human fine tuning.


\begin{center}
\includegraphics[width=\textwidth]{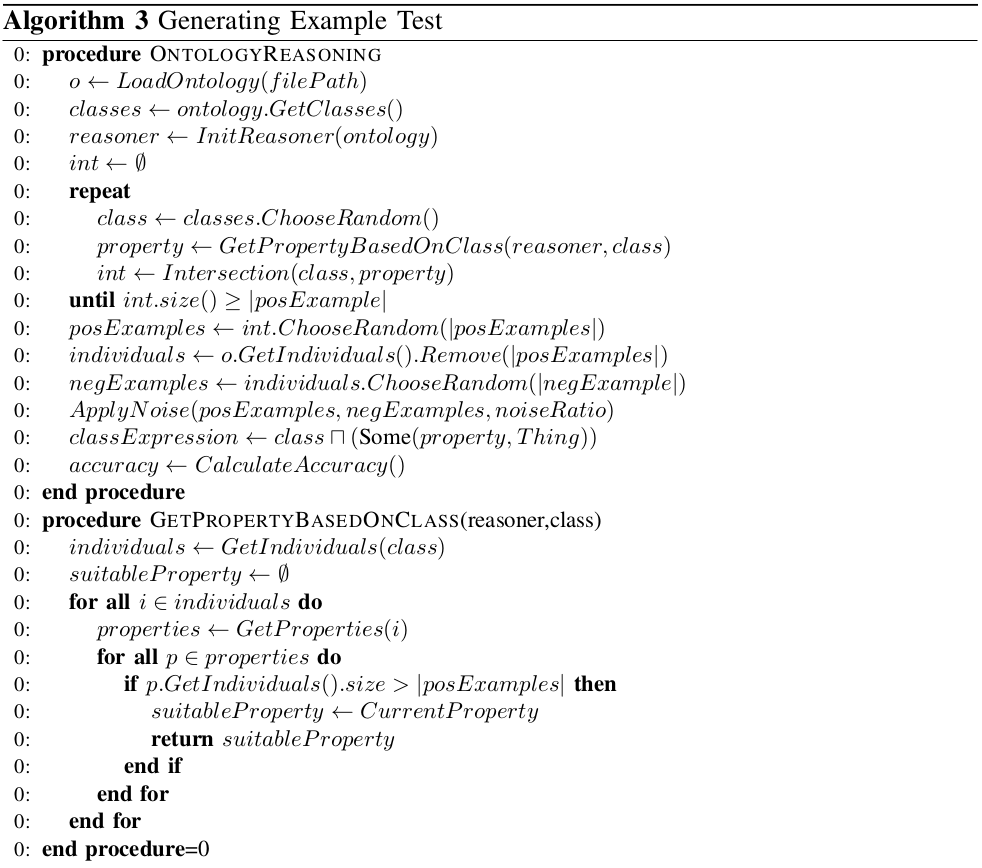} 
\end{center}

\subsection{Heuristics designed for FM approach}

The proposed $HT_1$ heuristic differs from the standard CELOE heuristic by considering the parent node's refinement count, which is essentially the number of its child nodes. 
The premise behind this heuristic is the potential value of less-branching paths in the search tree, which might prove beneficial in later steps.

In the standard CELOE heuristic, branches with more child nodes (or refinements) are prioritized, as they are often seen as more promising. 
However, $HT_1$ posits that less-branching paths (those with fewer child nodes) could also be of value. 
To encourage the exploration of these less-branching paths, $HT_1$ integrates an additional term into the final score calculation - the inverse of the parent's refinement count, multiplied by a weight factor. 
This effectively gives a score boost to nodes that have fewer siblings. 
\begin{center}
\begin{align}\label{eq:heuristic_test_1}
HT_1 = \begin{cases} 
start_{bonus} - (horiz - 1) \cdot \beta
- refin \cdot \gamma, & \text{if node = root} \\
(acc - acc_{parent}) \cdot \delta + 
\frac{1}{refin_{parent}} \cdot \epsilon 
- (horiz - 1) \cdot \beta - refin \cdot \gamma , & \text{otherwise}
\end{cases}\nonumber
\end{align}
\end{center}
Here, $start_{bonus}$ is score for the root of the search tree, 
$acc$ is the  accuracy of the evaluated node, while $acc_{parent}$ is the accuracy of the parent of the evaluated node. 
The $horiz$ parameter counts the number of horizontal expansions to reach the current node, or the length of the class expression. 
Number of refinements or the number of children in the search tree is represented by $refin$, while  
$refin_{parent}$ is the number of refinements of the parent, or the number of nodes on the same level that came from the same parent
The values $\alpha$, $\beta$, $\gamma$, $\delta$ are weights chosen based on the problem in the domain.

The $FH$ heuristic factors in both the depth of the current node within the search tree and its F1 score. 
The depth of a node in the search tree is the number of steps from the root to the node. Nodes deeper in the search tree often signify more "complex" solutions. 
Hence, we introduce a depth-based penalty to encourage simpler solutions in line with Occam's razor, the simplest solution is often the best, as shown in equation~\ref{eq:heuristic_F_1}:
\begin{equation}\label{eq:heuristic_F_1}
FH_1 = - horiz \cdot \alpha + 
\begin{cases} 
f1 \cdot \beta, & \text{if } f1 \geq 0.8 \\
    -f1 \cdot \gamma, & \text{if } f1 \leq 0.3 \\
0, & \text{otherwise}
\end{cases}
\end{equation}

Here, $horiz$ is depth of the node in the search tree, $\alpha$ is the penalty factor for the depth of a node in the search tree, $f1$ the F1 score of the node, 
$\beta$ the bonus factor for high F1 score (when $f1 \geq 0.8$), while  $\gamma$ is a penalty for very low F1 score (when $f1 \leq 0.3$).

\section{\textbf{Learning ontologies with forrest mixing approach}}
We illustrate the functionality of FM on a small sized ontology. 
The trace of FMA displays console outputs in small clusters, after which a brief explanation is provided. 
The ontology and examples were created manually in order to contain disjoint classes, in this case the classes $Student$ and $UniversityEmployee$ have common individuals. 
In the configuration file examples the individuals chosen as positive examples are all students, university employees and work in a research program. 
The goal of these examples is to better explain how the selection process of nodes work and how refinements are added to the search tree. 

The algorithm tries to find a class expression that best differentiates between the positive and negative training examples. 

\begin{small}
\begin{center}
\begin{lstlisting}[caption=Step 1: Identifying starting classes for search trees, label=lst:s1]
Running algorithm instance "alg" (FM)
FMA starting
Nb of tree roots to find: 2
Thing  cov 1.0, ResearchProgram cov0.0, Student cov1.0, University cov0.0, UniversityEmployee cov1.0,  Student  cov 1.0,  UniversityEmployee  cov 1.0
2 trees found with roots: [Student, UniversityEmployee]
\end{lstlisting}
\end{center}
\end{small}

Within the DL-Learner framework, the FM algorithm is initialized with a specific configuration. 
The user selects the number of trees to be used, which in this case is set to 2 (Listing~\ref{lst:s1}). 
FM then proceeds to generate and explore multiple classes, beginning with the top concept ($\top$), and iteratively specializes them until they cannot be further specialized without compromising the coverage of positive examples. 
The first two classes obtained through this process are identified as the best starting classes as the roots of our search trees.

\begin{small}
\begin{center}
\begin{lstlisting}[caption=Step 2: Picking the most promising class for refinement,label=lst:s2]
Student acc: 0.6
Best description so far: Student 
acc: 0.6 
f-score: 0.6666666666666666 
ref: 0 time: 9
UniversityEmployee acc: 0.6
\end{lstlisting}
\end{center}
\end{small}
Since $Student$ is the first node is selected as best one so far. 
The accuracy, f-score, number of refinements required to get the expression and time in ms are also displayed. (Listing~\ref{lst:s2}). 
\begin{small}
\begin{center}
\begin{lstlisting}[caption="Step 3: Refining the current class",label=lst:s3]
Node Student score calculation : 
    Horizontal expansion: 1.0
    Start node: 1.0
    Acc gain: -1.0
    Parent Refinements: 0.0
    Refinements: 0.0
    score: 0.7
CURRENT TREE WITH ROOT: Student
Current node: Student, accuracy: 0.6
Horizontal Expansion: 1
REF added from conj: 
Refinements for node Student: []
\end{lstlisting}
\end{center}
\end{small}

The best node from the tree is selected and its score calculation is displayed. Since the expansion is 1, we can not find refine a new class expression with length 1 (Listing~\ref{lst:s3}).
\begin{small}
\begin{center}
\begin{lstlisting}[caption=Step 4: Selecting another node for expansion,label=lst:s4]
Node UniversityEmployee score calculation : 
    Horizontal expansion: 1.0
    Start node: 1.0
    Acc gain: -1.0
    Parent Refinements: 0.0
    Refinements: 0.0
score: 0.7
REF added from conj: 
Refinements for node UniversityEmployee: []
\end{lstlisting}
\end{center}
\end{small}

After the first tree either has no refinements or added the maximum number of nodes, the second tree best node is selected (Listing~\ref{lst:s4}). 
Again, horizontal expansion is 1 and no refinements are found.
\begin{small}
\begin{center}
\begin{lstlisting}[caption=Step 5: Going back to the first tree to find expansions,label=lst:s5]
Node Student score calculation : 
    Horizontal expansion: 3.0
    ...
    score: 0.49999999999999994
CURRENT TREE WITH ROOT: Student
Current node: Student, accuracy: 0.6
Horizontal Expansion: 3
Refinements for node Student: [Student and Student, Student and UniversityEmployee]
Selected refinement: Student and Studentt acc: 0.6
    Node Added
Selected refinement: Student and UniversityEmployee: acc 0.6
    Best description so far: Student and UniversityEmployee acc: 0.8
    Node Added
\end{lstlisting}
\end{center}
\end{small}

We select the node $Student$ again, we are back to the first search tree, but this time the expansion is 3 and we find a refinement (Listing~\ref{lst:s5}). 
The best current expression for the target class is $Student \sqcap UniversityEmployee$.

\begin{small}
\begin{center}
\begin{lstlisting}[caption=Step 6: Using conjunction as refinement,label=lst:s6]
Node Student and UniversityEmployee score calculation : 
Horizontal expansion: 3.0
...
score: 0.6600000000000001
REF added from conj: 
(Student and UniversityEmployee)
Refinements for node Student and UniversityEmployee: []
Selected refinement: (Student and UniversityEmployee)
(Student and UniversityEmployee) acc: 0.8
Added node: (Student and UniversityEmployee)
\end{lstlisting}
\end{center}    
\end{small}

Listing~\ref{lst:s6} shows that nodes created from the conjunction ($\sqcap$) of best nodes form different trees are created. 
This node was already added in the tree with root $Student$, but here its first added as the conjunction before it is added as a normal refinement.

\begin{center}
\begin{small}
\begin{lstlisting}[caption=Step 7: Using quantified relations as refinement,label=lst:s7]
Node Student and UniversityEmployee and (not (ResearchProgram)) score calculation : 
Horizontal expansion: 6.0
...
CURRENT TREE WITH ROOT: UniversityEmployee
...
Student and UniversityEmployee and (inProgram some Thing) acc: 1.0
 Best description so far: Student and UniversityEmployee and (inProgram some Thing) acc: 1.0 f-score: 1.0 ref: 40 time: 71
Added node: Student and UniversityEmployee and (inProgram some Thing)
\end{lstlisting}
\end{small}
\end{center}

In Listing~\ref{lst:s7} the best class expression is found
\begin{equation}
Student  \sqcap  UniversityEmployee  \sqcap  (\exists inProgram.\top).     
\end{equation}

We can see the accuracy, f-score, the number of refinements needed to get here and time in miliseconds displayed. 
From this point onward the algorithm will search for better class expressions but it won't find any. 

\section{\textbf{Running Experiments}}
Before presenting the results, we briefly introduce what constitutes a test, the setup and algorithms employed.

\subsection{Experiments Setup}
A test in the context of ILP and hypothesis search can be defined as a triplet denoted by $(\mathcal{K}, E, \mathcal{C})$, where $\mathcal{K}$ represents the knowledge base, in our case an ontology in the OWL format, $E$ represents the set of examples, and $\mathcal{C}$ represents the target class expression. 
For testing the performance of FM, (1) we used datasets from the DL-Learner and additionally (2) we created our own synthetic datasets tailored to specific testing scenarios. 
For the knowledge base $\mathcal{K}$, we used the UOBM generator.

For $E$ and $\mathcal{C}$ in the test triplet, we designed a Java algorithm that finds a class expressions of the format $classA \sqcap \exists hasRelationR.Thing$. 
This class expression is found in previously generated ontology $\mathcal{K}$.
We chose this simple structure as the majority of relations we seek are simple. 
While this approach uses brute force, and may not be the most efficient, it serves our requirements due to the simplicity of the expressions.

The users can specify a minimum number of positive examples, 
After class expressions are found and positive and negative examples are determined, we add an additional layer of noise to our testing. 
We randomly remove 5\% of examples from both positive and negative sets, followed by a swapping of examples between the two sets. 
The swapping guarantees that the accuracy is not 1.0, because we rarely see this example in real life. 
The deletion ensures that the examples do not cover all individuals of the class, which is again not a real scenario.

\subsection{Results}
To test the Forest Mixing approach, we used two datasets: (i) one real-world dataset known as the Carcinogenesis dataset from the DL-Learner, and (ii) a synthetic dataset tailored to our specific testing scenarios. 
The Carcinogenesis dataset revolves around compounds and cancer-related data and contains 142 classes, 4 object properties, 15 data properties and 22.372 individuals. 
The synthetic dataset which we created consists of 40 classes, 6 object properties, 15 data properties and 26.766 individuals. 
These datasets were chosen for their ability to represent general cases, rather than specific ones where FMA is theoretically optimized.

The testing employed different heuristics as part of the FM approach. 
The use of these various heuristics aids in exploring the impact on performance and results under a broad array of scenarios, thereby providing a well-rounded understanding of FM's capabilities.
Table~\ref{table:carcinogenesis} presents the metric results and the corresponding class expressions learned from the carcinogenesis dataset, while Table~\ref{table:algUoBM} for the syntetic dataset.
Here,  FM1 represents the FM algorithm with a single search tree and FM2 represents the FMA algorithm with two search trees.

\begin{table*}
\centering
\caption{FM performance with various heuristics on the Carcinogenesis dataset}
\label{table:carcinogenesis}
\begin{tabular}{llllp{7cm}}
\hline
\textbf{Algorithm} & \textbf{Pred. Acc.} & \textbf{F1} & \textbf{Time (m)} & \textbf{Class expression} \\
\hline
OCEL & 0.68 & - & ~10 & 
$(\geq 2)hasBond.Bond \sqcap \exists  amesTestPositive.\top) \sqcup 
\exists hasStructure.Ar\_halide$\\
CELOE & 0.63 & 0.65 & ~10 & 
$\exists hasStructure.(Amino \sqcup Halide)$\\
NaiveALLearner & 0.53 & - & ~10 &  $\top$\\
FM1 + $HT_1$ & 0.61 & 0.61 & ~10 & 
$ Compound \sqcap \exists amesTestPositive.\top $\\
FM2 + $HT_1$ & 0.62 & 0.71 & ~50  &   
$Compound \sqcap (\leq 3) hasAtom.Carbon10$\\
FM + $FH$ & 0.61 & 0.61 & ~10 & 
$Compound \sqcap \exists amesTestPositive.\top$\\
\hline
\end{tabular}
\end{table*}

\begin{table*}
\centering
\caption{FM performance with various heuristics on the syntetic dataset}
\label{table:algUoBM}
\begin{tabular}{llllp{7.5cm}}
\hline
\textbf{Algorithm} & \textbf{Pred. Acc.} & \textbf{F1} & \textbf{Time (m)} & \textbf{Class expression} \\ \hline
OCEL & 0.68 & - & ~10 & 
$UndergraduateStudent \sqcup \lnot Woman \sqcap \lnot Article \sqcap \lnot Software \sqcap \lnot Specification$\\
CELOE & 0.69 & 0.76 & ~10 & 
$Book \sqcup \lnot AssociateProfessor \sqcap \forall publicationAuthor.Employee$\\
FM1 + $HT_1$ & 0.62 & 0.72 & ~10 & 
$(< 1) publicationAuthor. Employee$\\
FM2 + $HT_1$ & 0.69 & 0.76 & ~50 & 
$Book \sqcup \lnot Professor \sqcap \forall publicationAuthor.Employee$\\
FM + $FH$ & 0.62 & 0.72 & ~10 & 
$(< 1) publicationAuthor. Employee$\\
\hline
\end{tabular}
\end{table*}

Furthermore, we evaluated FM's performance in comparison to CELOE in scenarios involving non-disjoint classes. For this experimental context, we created a compact ontology encapsulating such non-disjoint classes. This model represents a small segment of a university ecosystem, comprising students, university employees, each of whom can be associated either with a research program, a university, or both. T
his ontology consists of 4 classes, 2 object properties, 0 data properties and 11 individuals.

Our underlying assumption here was that FM, due to its inherent design advantages when dealing with non-disjoint classes, should outperform CELOE in terms of finding the correct class expression in a more efficient manner.
The outcomes of these tests with corresponding class expressions are listed in Table~\ref{table:resultsFMACELOEStudent}. 
The first nine tests have learned the same class expression:
\begin{eqnarray}
Student \sqcap UniversityEmployee \sqcap 
\exists inProgram.ResearchProgram
\end{eqnarray}
The tenth approach, i.e. FMA2,  learned the distinct expression:
\begin{eqnarray}
Student \sqcap \exists inProgram.ResearchProgram \sqcap \nonumber \\ 
UniversityEmployee \sqcap \exists inProgram.\top
\end{eqnarray}

\begin{table}
\centering
\caption{Results CELOE and  FMA on non-disjoint Class Ontology}
\label{table:resultsFMACELOEStudent}
\begin{tabular}{llrr}
\hline
 \textbf{Algorithm} & \textbf{Starting Class} & \textbf{Time (ms)} & \textbf{Refinements} \\
\hline
CELOE & $\top$ & 133 & 88  \\
 CELOE & Student & 83 & 16  \\
 CELOE & UniversityEmployee & 86 & 16   \\
FMA1 & $\top$ & 41 & 28   \\
FMA1 & Student & 37 & 28  \\
 FMA1 & UniversityEmployee & 38 & 28  \\
 FMA2 & $\top$ & 44 & 42  \\
FMA2 & Student & 41 & 31  \\
FMA2 & UniversityEmployee & 41 & 32  \\
 FMA2 & Both & 53 & 42  \\
\hline
\end{tabular}
\end{table}

\begin{figure*}
\centerline{\includegraphics[width=1\textwidth]{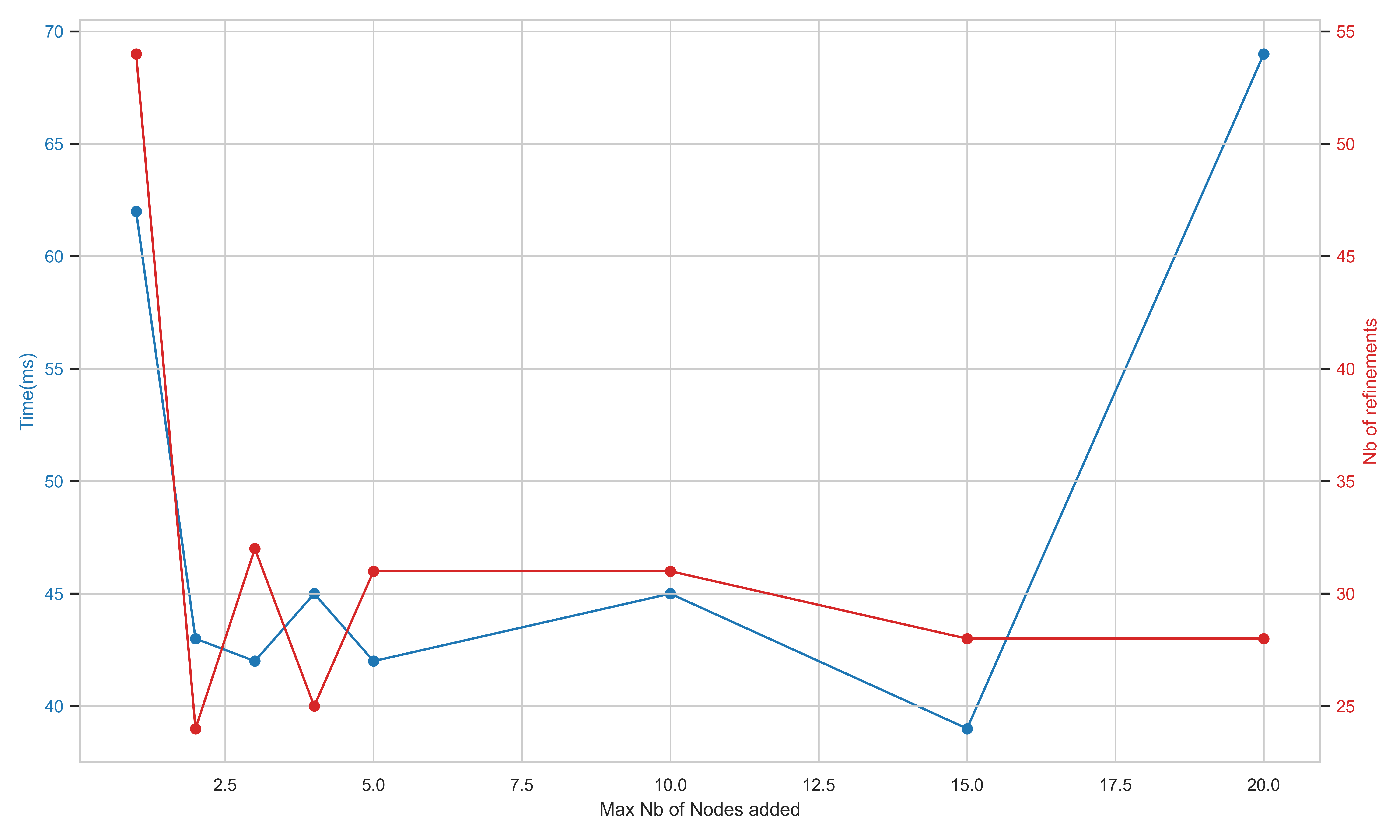}}
\caption{Comparison of Time and Number of Refinements Relative to Maximum Number of Nodes Added to a Search Tree}
\label{fig:NumberOfAddedNodes}
\end{figure*}

One additional test was conducted. 
FMA requires a parameter for limiting the number of nodes a search tree can it to itself at once. In the previous testing we noticed that the more trees we have the more refinements we need to find our class expression. 
In order to have a conclusion we isolated that case, using FMA with two search trees and $Student$ as starting class and we varied the number of nodes allowed for a tree to add. 
In Figure~\ref{fig:NumberOfAddedNodes} the x-axis denotes the parameter \texttt{maxNodesAddedPerTree}, which represents the maximum number of nodes a tree can incorporate into itself during a single cycle. The y-axis simultaneously tracks two different metrics, distinguished by color. The red line illustrates the changes in the number of refinement iterations required, while the blue line maps out the time consumed.

Our observations from the graph suggest that the addition of more than one node leads to a near-constant requirement of refinements. This insinuates that in a small ontology environment, the system behaves akin to a single tree with the inclusion of two or more nodes. The time consumption, depicted by the blue line, increases at the extreme points of \texttt{maxNodesAddedPerTree}, yet this parameter negligibly affects the system's overall performance due to minor temporal differences.

\section{\textbf{Conclusion}}
We examined the potential benefits of using multiple search trees and a shared pool of refinements as an enhancement to CELOE in the context of DL-Learner. 
Our initial hypothesis suggested that the Forest Mixing approach would outperform CELOE when handling non-disjoint classes and specific target class expressions.
The results from our experiments indicate that, contrary to our research hypothesis, FM is less efficient than CELOE. 
Furthermore, in the context of FM alone, an increase in the number of trees surprisingly appears to negatively impact performance. 
These findings prompt further investigation to fully understand the factors influencing the performance of FM and how it could potentially be optimized for the task at hand.

It is also conceivable that the current FM algorithm may not be fully optimized with regard to the number of refinements it uses in order to generate the target class expressions. 
Given that the number of refinements can substantially increase the complexity of the search space, excessive refinement operations may result in performance degradation.
Using different heuristics and refinement operators could potentially enhance the algorithm's performance, as the ones used in CELOE might not be the most suitable for this algorithm.

Additionally, a deeper investigation into the core logic of FM, particularly the management of the shared pool, could yield valuable insights. Experimenting with various types of pool management techniques and strategies for combining the most promising nodes from each tree could further refine the efficacy of the algorithm.

\section*{Acknowledgement}
A. Groza is supported by the project number PN-III-P2-2.1-PED-2021-2709, within PNCDI III. 


\begin{thebibliography}{00}

\bibitem{b1} L. Bühmann, J. Lehmann, and P. Westphal, "DL-Learner -- A framework for inductive learning on the Semantic Web," J. Web Semantics, vol. 39, pp. 15-24, 2016.

\bibitem{b2} J. Lehmann, S. Auer, L. Bühmann, S. Tramp, "Class expression learning for ontology engineering," J. Web Semantics, vol. 9, no. 1, pp. 71-81, 2011.

\bibitem{b3} A. Srinivasan, "The Aleph Manual", [Online]. http://web.comlab.ox.ac.uk/oucl/research/areas/machlearn/Aleph/, 2003.

\bibitem{b9} P. Mateiu, A. Groza, and C. Nica. "Learning Ontologies with Relational Concept Analysis." 2022 IEEE 20th Jubilee World Symposium on Applied Machine Intelligence and Informatics (SAMI). IEEE, 2022.

\bibitem{b10} P. Mateiu and A. Groza. "Ontology engineering with Large Language Models." in  \textit{25th International Symposium on Symbolic and Numeric Algorithms for Scientific Computing (SYNASC23), Nancy, France}, 11-14 September 2023, arXiv preprint arXiv:2307.16699 (2023).


\bibitem{b20} Marginean, Anca Nicoleta, and Kando Eniko. "Towards lexicalization of dbpedia ontology with unsupervised learning and semantic role labeling." 2016 18th International Symposium on Symbolic and Numeric Algorithms for Scientific Computing (SYNASC). IEEE, 2016.


\bibitem{b4} F. M. Donini, "Complexity of Reasoning," in \textit{The Description Logic Handbook: Theory, Implementation and Applications}, F. Baader et al., Eds. Cambridge University Press, 2003.

\bibitem{b5} D. Calvanese and G. De Giacomo, "Expressive Description Logics," in \textit{The Description Logic Handbook: Theory, Implementation and Applications}, F. Baader et al., Eds. Cambridge University Press, 2003.

\bibitem{b6} I. Horrocks, "Implementation and Optimisation Techniques," in \textit{The Description Logic Handbook: Theory, Implementation, and Applications}, F. Baader et al., Eds. Cambridge University Press, 2003.


\bibitem{b7} L. Badea and S. Nienhuys-Cheng, "A refinement operator for description logics," in \textit{International Conference on Inductive Logic Programming}, Springer, pp. 40-59, 2000.

\bibitem{b8} Y. Zhou, B. C. Grau, and I. Horrocks, "UOBM Generator," Available at: \url{https://www.cs.ox.ac.uk/isg/tools/UOBMGenerator/}.


\end{thebibliography}

\vspace{1cm}
\textbf{Authors}
\vspace{1cm}

\begin{minipage}{0.68\linewidth}
Marco Pop-Mihali has graduated from Technical University of Cluj-Napoca, Romania, Department of Computer Science. His research interests are related to artificial intelligence, currently working at Bosch as a data scientist, solving problems related to time series prediction, augmenting input data for large models, and processing big data.
\end{minipage}
\hfill
\begin{minipage}{0.26\linewidth}
\includegraphics[width=1\linewidth]{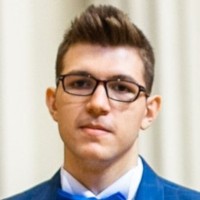}
   
\end{minipage}
\vspace{1cm}

\begin{minipage}{0.68\linewidth}
\href{https://users.utcluj.ro/~agroza/publications/}{Adrian Groza} is professor of Artificial Intelligence at Technical University of Cluj-Napoca, Romania, Department of Computer Science. His current research is related to knowledge representation and reasoning. He recently published the book \href{https://users.utcluj.ro/~agroza/puzzles/maloga/slides.html}{Modelling puzzles in first order logic}, Springer, that provides a collection of warm-up and fun activities to start a lecture on logic or computer science.
\end{minipage}
\hfill
\begin{minipage}{0.26\linewidth}
  \includegraphics[width=1\linewidth]{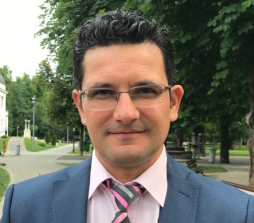}
    
\end{minipage}

\end{document}